\documentclass{article}
\usepackage{arxiv}

\usepackage[utf8]{inputenc} 
\usepackage[T1]{fontenc}    
\usepackage{hyperref}       
\usepackage{url}            
\usepackage{booktabs}       
\usepackage{amsfonts}       
\usepackage{nicefrac}       
\usepackage{microtype}      
\usepackage{lipsum}		
\usepackage{graphicx}
\usepackage[numbers]{natbib}
\usepackage{doi}

\usepackage{amsmath}
\usepackage{mathtools}
\usepackage{xcolor}
\usepackage{comment}
\usepackage{makecell}
\usepackage{multirow}
\usepackage{changepage}
\usepackage{hyperref}
\usepackage{caption}
\usepackage{footnote}
\usepackage{arydshln}
\usepackage{soul}

\usepackage{pifont}
\newcommand{\cmark}{\ding{51}}%
\newcommand{\xmark}{\ding{55}}%

\title{SoK: Verifiable Cross-Silo FL}

\date{September 9, 2024}	

\author{ \href{https://www.linkedin.com/in/aleksei-korneev-3927aa218/}{\hspace{1mm}Aleksei Korneev} \\
	University of Lille, Inria, \\ CNRS,
        Centrale Lille,  \\ UMR 9189 - CRIStAL \\ 
	\texttt{aleksei.korneev@inria.fr} \\
	\And
	{Jan Ramon} \\
	University of Lille, Inria, \\ CNRS,
        Centrale Lille,  \\ UMR 9189 - CRIStAL \\ 
	\texttt{jan.ramon@inria.fr} \\
}

\renewcommand{\shorttitle}{Verifiable Cross-Silo FL}

\hypersetup{
pdftitle={SoK: Verifiable Cross-Silo FL},
pdfauthor={Aleksei Korneev, Jan Ramon},
pdfkeywords={federated learning, FL, cross-silo FL, verification, verifiable protocols, zero knowledge proofs, ZKP},
}

\begin{document}
\maketitle

\begin{abstract}
    Federated Learning (FL) is a widespread approach that allows training machine learning (ML) models with data distributed across multiple devices. In cross-silo FL, which often appears in domains like healthcare or finance, the number of participants is moderate, and each party typically represents a well-known organization. For instance, in medicine data owners are often hospitals or data hubs which are well-established entities.  However, malicious parties may still attempt to disturb the training procedure in order to obtain certain benefits, for example, a biased result or a reduction in computational load. While one can easily detect a malicious agent when data used for training is public, the problem becomes much more acute when it is necessary to maintain the privacy of the training dataset. To address this issue, there is recently growing interest in developing verifiable protocols, where one can check that parties do not deviate from the training procedure and perform computations correctly. In this paper, we present a systematization of knowledge on verifiable cross-silo FL. We analyze various protocols, fit them in a taxonomy, and compare their efficiency and threat models. We also analyze Zero-Knowledge Proof (ZKP) schemes and discuss how their overall cost in a FL context can be minimized. Lastly, we identify research gaps and discuss potential directions for future scientific work.

\end{abstract}


\section{Introduction}

Nowadays, the broad propagation of ML technologies is rapidly increasing. ML applications affect a variety of fields such as medicine, finance, marketing, education, and many others \cite{MedicineMLSurvey, FinanceMLSurvey, MarketingMLSurvey, EducationMLSurvey}. Many ML approaches rely on a process of training on historical data: a model learns statistical patterns that later allow new predictions to be inferred. However, in some cases, data may contain private or confidential information; therefore, access to such data is limited, and applying ML must be done with extreme caution, either due to an interest in privacy of the data owners (DOs), e.g., individual persons caring about their privacy or companies caring about intellectual property, or for regulatory compliance, e.g., with the General Data Protection Regulations (GDPR).
 
In FL, multiple DOs, who are also sometimes referred to as clients, can train a model together, possibly under coordination of a central server, by exchanging encrypted messages without revealing their private data. As a result, researchers can benefit from a large amount of shared data and at the same time preserve privacy. However, while preserving privacy in FL allows protecting sensitive information, at the same time it produces an additional challenge in the verification of participants' behavior. Indeed, due to a possibility of malicious actions, it is important to ensure that all calculations are performed correctly even if the used data is private.

FL is often divided into two categories: cross-device and cross-silo. In the cross-device FL setting, data comes from a large number of small and usually anonymous devices with low computational capacities. Anonymity complicates penalizing clients; a single client is free to abort or to violate the procedure at any time. In contrast, in this paper we focus on cross-silo FL where the number of parties is moderate; each party is usually a well-known and large entity that is expected to cooperate in the entire training process via devices with high computing power. Each party has an incentive to care about its reputation and can be held liable if it is found to be fraudulent. For instance, cross-silo FL appears in the healthcare domain, where DOs are medical centers or hospitals that collaborate to train ML models to improve patient care. Although the need for countermeasures against malicious attacks in such setting could be reduced due to the liability of participants, verification of the calculations is still required to establish confidence in FL’s performance.

Recently, dozens of research works devoted to verifiable FL have been published, proposing methods to ensure the verifiability of the parties' computations, using different infrastructures and relying on various assumptions. Nonetheless, to the best of our knowledge, verifiability in the context of cross-silo FL has not been thoroughly studied. Chao et al. in \cite{CSFLSurvey} studied challenges of cross-silo FL setting in details, but the verifiability property was not taken into account. In  \cite{TowardsVerifiableFL, TrustworthyFL} authors were focused on verifiability in FL, however features of the cross-silo setting were not considered and protocols' efficiency was not analysed. Mansouri et al. presented a SoK paper \cite{SAPSOK} devoted to secure aggregation protocols and included verification in the list of challenges, nevertheless, specific features of cross-silo FL were not in the scope of the paper and an efficiency analysis of verification techniques was not performed. Lastly, in \cite{ModulusSurvey, ZKPmeetsML} authors studied applications of various ZKP schemes for ML, but these works do not address FL.

In this paper, we provide a systematization of knowledge on verifiable cross-silo FL. Our contributions are summarized as follows:

\begin{itemize}
    \item to the best of our knowledge, we are the first to conduct a survey on verifiable FL while studying specific challenges of the cross-silo setting;
    \item we propose a new taxonomy of existing verifiable cross-silo FL protocols while analyzing their efficiency and threat models;
    \item we perform a comparison of ZKP schemes from the perspective of applying them in cross-silo FL and discuss the influence of parameters such as the ZKP scheme and circuit size on the cost;
    \item we define future challenges and identify research gaps.
\end{itemize}

The rest of the paper is organized in the following way: Section 2 introduces some background and notations of FL and crpytographic primitives; Section 3 presents an analysis of existing verifiable cross-silo FL protocols; Section 4 is devoted to a comparison of ZKP schemes from the perspective of their applicability for cross-silo FL, then we describe a storage cost optimization for ZKP-based FL protocols; Section 6 describes challenges and research gaps; and Section 7 concludes the paper.

\section{Background}

\subsection{FL process}

The majority of FL approaches contain two main categories of operations: local operations (both at the side of the clients and the server) and aggregation of clients' values. Additionally, there are also other operations specific to certain FL protocols, for example, where participants should draw random numbers, exchange cryptographic keys or select a subset of parties to communicate with.

In this paper, we consider both settings where the aggregation is coordinated or performed by a central server and settings where DOs perform the aggregation in a decentralized way.

Some ML algorithms involve running an optimization algorithm, e.g., stochastic gradient descent (SGD).  We refer to each iteration of such algorithm as an epoch.

\subsection{Cross-silo FL properties}
While analyzing the suitability of various algorithms for the cross-silo FL setting, we assume that: 

\begin{enumerate}
    \item the number of participants is moderate (at most several thousands);
    \item all participants have an incentive to care about their reputation, they may only cheat in a way which can not be detected by others;
    \item all participants agree on the model to be trained (type of calculations to be executed);
    \item DOs possess computationally sufficiently powerful equipment;     
\end{enumerate}

\subsection{Adversarial attacks on FL}

In a survey on FL threats by Rodríguez-Barroso et al. \cite{FLattacksSurvey} two classes of adversarial attacks were distinguished:
\begin{itemize}
\item privacy attacks, whose purpose is to infer sensitive information from the learning process;
\item attacks which aim at modifying the behaviour or output of the FL process.
\end{itemize}

The development of a FL protocol that is resistant to adversarial attacks requires applying a combination of various privacy enhancing technologies (PETs). For example, in order to prevent privacy attacks, authors of state-of-the-art solutions employ differential privacy (DP), multi-party computation (MPC), secure shuffling and Trusted-Execution Environment (TEE)  among others.

In this paper, we study verification techniques that allow mitigating the second class of attacks, attacks on the federated learning process, such as data or model poisoning, when an adversary intentionally uses incorrect data or performs computations incorrectly to bias the resulting model. Authors of the considered papers applied commitment schemes, homomorphic hash functions, ZKP schemes and other methods to ensure that the federated model is computed correctly. A detailed analysis of these methods is presented in sections \ref{ApproachesAnalysis} and \ref{ZKPschemesAnalysis}.

\subsection{Verifiable FL}

In the scope of this paper, we rely on a definition of Verifiable FL inspired by \cite{TowardsVerifiableFL}:

\label{VerifFLDefinition} 
\textit{Definition (Verifiable FL).} FL is verifiable if selected parties are able to verify that the tasks of all participants are correctly performed without deviation.

Following this definition, in contrast to the survey \cite{TrustworthyFL}, we only include approaches that at least partly verify computations of the FL process. For example, we do not analyze protocols which are focused only on verification of identity, ownership, or data provenance. We also exclude protocols considered in \cite{SAPSOK} that aim to prevent model poisoning attacks by analyzing distribution of values submitted by parties. Such methods efficiently mitigate some attacks, but do not allow to verify the correctness of individual computations or of the individual uses of the input data, e.g., an individual outlier input value is infrequent but possibly valid. Moreover, their efficiency depends on  the domain and an attacker strength. On the other hand, we do include in our analysis several protocols devoted to verifiable federated private averaging and verifiable cross-device FL since the same verification techniques could be used in the cross-silo FL setting.

\subsection{Threat models}
\label{threat_models}
In the scope of the considered works, authors usually rely on two widely-spread types of threat models: honest-but-curious (a.k.a. semi-honest) and malicious. According to the standard cryptography definitions, an \textit{honest-but-curious} agent does not deviate from the protocol, but keeps a record of the protocol transcript and analyzes it to gain information about other users, while a \textit{malicious} adversary can deviate from the prescribed protocol instructions and follow an arbitrary strategy to obtain greater benefits. However, in the context of FL, authors often adapt these definitions with additional properties. In order to thoroughly analyze miscellaneous flavors of the applied threat models we distinguish the following four categories:  

\begin{itemize}
    \item \textbf{honest (or trusted)}: always follows the protocol correctly and is trusted with sensitive information;
    \item \textbf{honest-but-curious}: always follows the protocol correctly, but is not trusted with sensitive information;
    \item \textbf{forger}: may try to forge different data, but otherwise follows the protocol, is not trusted with sensitive information;
    \item \textbf{malicious}: can arbitrary deviate from the protocol and is not trusted with sensitive information.
\end{itemize}

In the scope of this paper, in order to describe different approaches in a rigorous manner, we specify the robustness of protocols to participant drop-outs, i.e. agents who register to participate but subsequently abandon the protocol, separately from the aforementioned categories of threat models. For example, to report that a protocol is robust against forging and drop-outs, we denote its threat model as "forger + drop".

\subsection{Blockchain technology}

A blockchain is a sequence of blocks, which holds a complete list of transaction records like a conventional public ledger \cite{BC_definition}. A transaction could be any action taking place on a blockchain network, for example, a transfer of digital currency from one party to another. In order to achieve the agreement about the ledger's state, blockchains rely on consensus mechanisms. For instance, one of the most popular mechanisms is a Proof of work (PoW), where parties, called miners, calculate a hash of the constantly changing block header. When one miner obtains a relevant value, all others must confirm its correctness and add a collection of transactions used for the calculations as a new block. 

In certain blockchain settings, parties can deploy on the blockchain code scripts, called smart contracts, that run synchronously on multiple nodes of the blockchain. Smart contracts can be stored in the blockchain and can be automatically executed when certain pre-conditions are met \cite{BC_smartcontracts}.

\subsection{Commitment scheme}

A commitment scheme (CS) is a cryptographic primitive that allows parties to commit to values while keeping them hidden from others \cite{FirstCS}. A party cannot modify the value after committing to it, but can later reveal it. CSs have two properties: hiding, meaning that a commitment reveals nothing about the original value, and binding, meaning that a party cannot compute the same commitment from different values, typically due to a computationally hard underlying problem. Lastly, some CSs are also homomorphic, meaning that there are two binary operations $+$ and $\cdot$ defined in the domain of original values and the domain of their commitments respectively, such that the following condition holds: $C(a+b) = C(a) \cdot C(b)$, where $a, b$ are values possessed by a party (or parties) and the commitment is represented by $C$.

\subsection{Zero knowledge proof}

A Zero-Knowledge Proof (ZKP) is a cryptographic method by which a party called the prover convinces another party, the verifier, about a statement over committed values \cite{FirstZKP}. A ZKP of a statement should satisfy three properties: 
\begin{itemize}
    \item completeness: if the statement is true, then an honest verifier will be convinced by an honest prover;
    \item soundness: if the statement is false, an honest verifier will be convinced with at most a negligible probability;
    \item zero-knowledge: if the statement is true, a verifier is not capable of learning anything but the proven statement itself.
\end{itemize}

Some ZKP can be turned into non-interactive proofs by replacing the verifier with a random oracle using the Fiat-Shamir heurisitic \cite{FiatShamir}. In the class of non-interactive ZKP, Zero Knowledge Succinct Non-interactive Arguments of Knowledge (zk-SNARKs) are of a particular interest due to the compact proofs relative to the size of the statement and fast verification \cite{IntroZKP}.

In FL, ZKPs can be used to prove that a party correctly performed prescribed computations and, therefore, did not deviate from the FL protocol.

\section{Verifiable cross-silo FL protocols analysis} \label{ApproachesAnalysis}

In this section, we present a taxonomy of existing verifiable cross-silo FL protocols, analyze the efficiency of verification techniques and threat models, and discuss the impact of the cross-silo setting on verification. In the scope of this section, we refer to the number of clients as $C$ and to the number of the aggregated vector dimensions as $D$. 

\begin{figure}[ht]
\centering
\includegraphics[width=\linewidth]{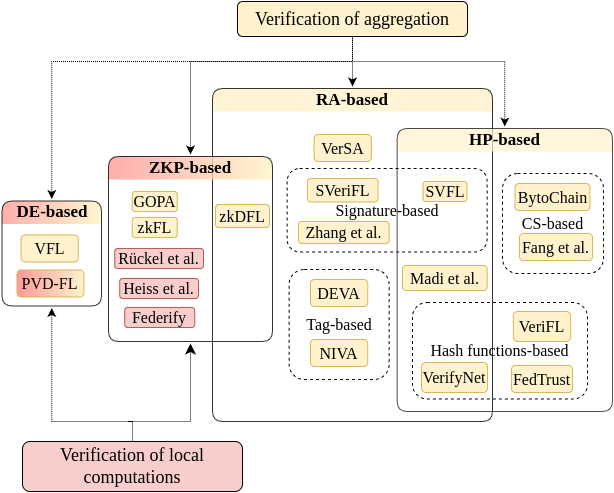}
\caption{A taxonomy of verifiable cross-silo FL protocols. The red color corresponds to approaches focused on the verification of clients' computations, the yellow color is used for approaches focused on the aggregation verification.} 
\label{FigureTaxon}
\end{figure}

In order to ensure that a FL protocol is executed correctly, one has to verify both the local computations of the clients, the aggregation, and the local computations of the server(s). We distinguish four categories of different verification techniques and describe each of them in details below. The full taxonomy is presented in Figure \ref{FigureTaxon}. Although each approach has specific characteristics, our categories allow observing general design patterns and infer conclusions about their efficiency. For this purpose, we assess computational and communication costs both per client and per server for each method. The comparison of asymptotic complexities of protocols and applied threat models is presented in Table \ref{VerAggTable} for approaches focused on verifiable aggregation and in Table \ref{VerLocModTable} for approaches focused on verification of local computations. 
We emphasize that complexity metrics are calculated specifically for the verification overhead and do not reflect computation and communication which is needed even if no verification is performed. Lastly, we assume that public key infrastructure, ML model weights and seeds of PRGs are initialized before the training procedure and do not require a presence of a trusted party. 

In order to fairly compare threat models of different approaches, we mapped the threat models described in the considered papers to definitions from Subsection \ref{threat_models} and assigned a suitable model ourselves in cases where authors did not explicitly describe it. In such cases, in tables \ref{VerAggTable} and \ref{VerLocModTable}, we use square brackets to denote assigned threat models and collusion markers. Additionally, some articles provided descriptions of multiple threat models, for example, separate models from privacy and verification perspectives; our table reflects only those related to verification. We note that assigning a threat model for approaches based on blockchain infrastructure poses additional challenges. While clients' computations in such settings are performed by the clients themselves, aggregation typically occurs as a smart contract. As a result, all miners redundantly compute the aggregation result and a model for the approaches that describe how to obtain it from the aggregation result. In such cases, in Tables \ref{VerAggTable} and \ref{VerLocModTable}, the server's threat model corresponds to the threat model of miners. However, it is important to mention that blockchain infrastructure itself is also vulnerable to specific threats, which are often not detailed in articles devoted to blockchain-based FL. As an example, Fang et al. \cite{SFCV-FL} explicitly mention that the protocol is robust when 70\% of the stake in the system is honest, but for others such assumptions are not always easily visible. In tables \ref{VerAggTable} and \ref{VerLocModTable}, the symbol "*" corresponds to a threat model applied to a fraction of agents. To sum up, threat models described in our tables for blockchain-based approaches are meaningful only when the assumptions of blockchain infrastructure itself are satisfied.

In tables \ref{VerAggTable}, \ref{VerLocModTable} we also mark whether methods are robust against client drop-out and/or server drop-out. For non-interactive methods (marked with "drop$^\dagger$"), clients only interact once to submit their contribution, which results in a rather trivial robustness against client drop-out: either a client submits the contribution and a later drop-out of the client is not relevant, or the client doesn't submit the contribution and then doesn't participate at all. For such approaches, once a client participates, his contribution will be taken into account in the computed aggregate if the protocol finishes. For interactive protocols, in all cases clients need to contribute to some form of "decryption" of the aggregate.  Some approaches \cite{GOPA} try to remove the input of the client(s) who dropped out so the set of clients who decrypt is exactly the set of clients who contributed input.  In other approaches \cite{VerifyNet, VerSA, DEVA}, robustness against client drop-out is achieved by a threshold secret sharing where only part of the clients are needed for this decryption.  This decreases security but avoids the need for computations to be rolled back. When we mark the server threat model as robust against drop out, in all cases there are multiple servers (in the case of blockchain based methods multiple miners) of which a certain fraction can drop out without preventing the computation of the final result, e.g., due to a threshold secret sharing scheme among servers or due to the consensus properties of blockchains.

\begin{table*}[]
\begin{adjustwidth}{-1cm}{}
    \begin{tabular}{|c|c|c|c|c|c|c|c|c|c|}
        \hline
        \multirow{2}{*}{\textbf{Approach}} & \multicolumn{2}{c|}{\textbf{Computational cost}} & \multicolumn{2}{c|}{\textbf{Communication cost}} &  \multicolumn{2}{c|}{\textbf{Threat model}} & \multirow{2}{*}{\textbf{\makecell{S-C \\ col.}}}  & \multirow{2}{*}{\textbf{TA}} & \multirow{2}{*}{\textbf{BC}}  \\
          & \textbf{client} & \textbf{server} & \textbf{client} & \textbf{server} & \textbf{client} & \textbf{server} & & & \\ \hline
        
VerSA \cite{VerSA} & $O(D)$ & $O(CD)$ & $O(D)$ & $O(CD)$ & h-b-c + drop &  [forger] & \xmark & \xmark & \xmark \\ 
         SVeriFL \cite{SVeriFL} & $O(D)$ & $O(CD)$ & $O(D)$ & $O(CD)$ & h-b-c & forger & \xmark & \cmark & \xmark   \\
         Zhang et al. \cite{Zhangetal} & $O(D)$ & $O(C)$  & $O(1)$  & $O(C)$ & [hon] & forger  & [\xmark] & \xmark & \xmark \\
         DEVA \cite{DEVA} & $O(CD)$ & $O(CD)$ & $O(CD)$ & $O(CD)$ & h-b-c + drop & forger + drop  & \xmark & \xmark & \xmark \\
         NIVA \cite{NIVA} & $O(CD)$ &  $O(CD)$ & $O(CD)$ & $O(CD)$ & [h-b-c + drop$\dagger$] & [forger + drop] & [\xmark] & \xmark & \xmark\\ \hline
         
         SVFL \cite{SVFL} & $O(D)$ & $O(C)$ & $O(1)$ & $O(C)$ & h-b-c  & [forger] & \xmark & \xmark & \xmark \\
         Madi et al. \cite{SFCV-FL} & $O(D)$ & $O(C)$ & $O(1)$ & $O(D)$ & hon & [forger] &  \xmark & \xmark & \xmark \\
         VerifyNet \cite{VerifyNet} & $O(D)$ & $O(CD)$ & $O(D)$ & $O(CD)$ & h-b-c + drop & forger & \xmark & \xmark & \xmark \\ \hline
         
         BytoChain \cite{BFLbasedBytoChain} & $O(C+D)$ & $O(D)$ & $O(1)$ & $O(1)$ & [mal + drop$\dagger$]  & [forger* + drop] & [\xmark] & \xmark & \cmark \\
         Fang et al. \cite{PPV-BC} & $O(C+D)$ & $O(1)$ & $O(1)$ & $O(1)$ & [hon + drop$\dagger$] & [forger* + drop] & [\xmark] & \xmark & \cmark  \\
         VeriFL \cite{VeriFL} & $O(C + \frac{D}{E})$ & $O(1)$ & $O(C)$ & $O(1)$  & h-b-c + drop  & forger & [\xmark] & CRS  & \xmark \\
         FedTrust \cite{FedTrust} &  $O(CD)$ & $O(1)$ & $O(CD)$ & $O(1)$  & hon & forger + drop & [\xmark] &  \xmark & \xmark \\  \hline

         zkDFL \cite{zkDFL} & $O(D)$ & $O(CD)$  & $O(C)$ & $O(C)$  & [hon + drop$\dagger$] & [forger*+drop] & [\xmark] & CRS & \cmark \\  \hline
         
         GOPA \cite{GOPA} & $O(D$log$C)$ & N/A &  $O(D$log$C)$ & N/A & forger* + drop  & N/A & N/A & \xmark & \xmark \\
         zkFL \cite{zkFL}  & $O(CD)$ & $O(CD)$ & $O(1)$ &  $O(C$log$(CD))$  & [hon + drop$\dagger$]& [forger+drop] &  [\xmark] & \xmark & 
         \xmark/\cmark \\ \hline
         VFL \cite{VFL}  &  $O(D)$  & $O(1)$ & $O(1)$ & $O(1)$ & h-b-c  & [forger] & [\xmark] & \xmark & \xmark \\
         PVD-FL \cite{PVD_FL} & $O(D)$ & N/A & $O(D)$ & N/A & [forger*] & N/A & N/A & \xmark & \xmark  \\
         \hline
    \end{tabular}
    \captionsetup{justification=centering}
    \caption{A comparison of asymptotic complexities of the aggregation verification overhead per epoch and threat models of verifiable FL protocols. Sections correspond to taxonomy categories (Figure \ref{FigureTaxon}). Notations: $E$ -- number of epochs, $C$ -- number of clients, $D$ -- number of vector dimensions, BC -- requires a blockchain infrastructure (the (significant) cost of the blockchain is not included in the table), S-C col. -- Server-Client collusion to bypass the verification, $[\ldots]$ -- not explicitly described in the paper.}
    \label{VerAggTable}
    \end{adjustwidth}
\end{table*}

\begin{table*}[]
\begin{adjustwidth}{-1cm}{}
    \begin{tabular}{|c|c|c|c|c|c|c|c|c|c|}
        \hline
        \multirow{2}{*}{\textbf{Approach}} & \multicolumn{2}{c|}{\textbf{Computational cost}} & \multicolumn{2}{c|}{\textbf{Communication cost}} &  \multicolumn{2}{c|}{\textbf{Threat model}} & \multirow{2}{*}{\textbf{\makecell{S-C \\ col.}}}  & \multirow{2}{*}{\textbf{TA}} & \multirow{2}{*}{\textbf{BC}} \\
          & \textbf{client} & \textbf{server} & \textbf{client} & \textbf{server} & \textbf{client} & \textbf{server} & & & \\ \hline
         R\"uckel et al. \cite{Ruckeletal} & O(D) & O(1) & O(1) & O(1)  & [forger + drop$\dagger$] & [forger* + drop] & \xmark  & CRS  & \cmark  \\ 
         Heiss et al. \cite{Heissetal} & O(D) & O(1) & O(1) & O(1) & [forger + drop$\dagger$] & [forger* + drop] & \xmark & CRS  & \cmark  \\ 
         Federify \cite{Federify} & O(D) & O(1) & O(1) & O(1)  & [forger + drop$\dagger$] & [forger* + drop] & \xmark & CRS  & \cmark  \\ 
         \hline
         PVD-FL \cite{PVD_FL} & $O(D)$ & N/A & $O(D)$ & N/A & [forger*] & N/A & N/A & \xmark  & \xmark  \\
         \hline
    \end{tabular}
    \captionsetup{justification=centering}
    \caption{A comparison of asymptotic complexities of the local computations verification overhead per epoch and threat models of verifiable FL protocols. Sections correspond to taxonomy categories (Figure \ref{FigureTaxon}). Notations: $C$ -- number of clients, $D$ -- number of vector dimensions, BC -- requires a blockchain infrastructure (the (significant) cost of the blockchain is not included in the table), S-C col. -- Server-Client collusion to bypass the verification, $[\ldots]$ -- not explicitly described in the paper.}
    \label{VerLocModTable}
    \end{adjustwidth}
\end{table*}

\subsection{Taxonomy description}

\textbf{Redundant aggregation (RA) based verification.} This category consists of approaches that require the server to aggregate some redundant values in order to prove that the aggregation of clients' values is performed correctly. This feature leads to a computational cost of the server to be at least $O(C)$. Moreover, some protocols \cite{DEVA, NIVA, VerSA} are designed under the assumption that each party has one secret value, therefore a naive scaling of the approach to a FL setting where parties share multi dimensional data would lead to an additional factor $D$ in the complexity.

In \cite{Zhangetal, SVeriFL} authors proposed to check that the result of aggregation is correct by means of cryptographic signature schemes based on bilinear pairings. In both works, the server has to compute a redundant aggregation of signatures. Indeed, such schemes allow ensuring that the aggregation result is obtained from data signed by all other clients, however a malicious server may aggregate arbitrary signed values (e.g., clients' values from previous epochs) and successfully pass the verification with a fabricated aggregated value. As a result, by relaxing the threat model, \cite{Zhangetal} achieves a better complexity than other protocols of the same category. 

In DEVA \cite{DEVA} and NIVA \cite{NIVA} decentralized FL protocols are proposed, all clients compute tags corresponding to the secret clients' values which are later used to verify that the aggregation is correct. Both protocols have an additional $D$ complexity factor due to the assumption that each client has only one secret value. In contrast to many others, DEVA and NIVA present solutions for a setting with multiple servers. Another RA-based approach is VerSA \cite{VerSA}. In VerSA all clients use the same secret vectors to compute a new input from their secret values and later check the consistency of the aggregated result with the aggregation of new inputs. Interestingly, authors provide a comparison with VerifyNet \cite{VerifyNet} approach (which lies in the intersection of RA- and HP-based categories) showing that despite the same verification asymptotic complexity, the VerSA's cost is orders of magnitude smaller.

Finally, from the treat model perspective, all RA-based approaches cope with a forger server while considering clients to be honest or honest-but-curious. Additionally, in \cite{SVeriFL}, integrity of data shared by clients is verified, this threat is processed by addition of a trusted authority (TA).

\textbf{Homomorphic property (HP) based verification.} This category covers verification techniques which rely on the HP of underlying primitives: hash functions \cite{VeriFL, FedTrust} and commitment schemes \cite{BFLbasedBytoChain, PPV-BC}. The general idea of such protocols is the following: clients compute hashes/commitments from their data and share them with each other, then all clients may verify the result of aggregation by checking that this result corresponds to the aggregation of hashes/commitments through homomorphism. As a consequence, both computational and communication costs of the server are $O(1)$ if the ciphertext length does not depend on $C$ or $D$. In order to verify the aggregation of clients' values, clients have to compute a hash/commitment in $O(D)$ from their data and to aggregate hashes/commitments from other clients in $O(C)$. In blockchain-based protocols clients upload hashes/commitments with a constant communication cost (to be multiplied with costs related to the blockchain infrastructure, see
Section \ref{blockchain_discussion}), while in other protocols, where clients have to exchange messages with each other, the communication cost per client is at least $O(C)$. Exceptionally, in FedTrust \cite{FedTrust} client costs have additional complexity factor $D$, as far as the hash is calculated for each component separately.

In VeriFL \cite{VeriFL}, apart from the main protocol, authors also demonstrated a way to optimize the verification. Leveraging the homomorphic property of the hash scheme and the repetitive nature of FL calculations, authors proposed an amortized verification which allows decreasing the number of hash function calls. Instead of checking that the product of hashes obtained from all clients is equal to the hash of the aggregated value obtained from the server after each epoch, one can draw a set of random coefficients to compute a linear combination of hashes obtained from all clients across multiple epochs and compare it with a hash of a linear combination of aggregated values obtained from the server using the same coefficients. As a result, in Table \ref{VerAggTable}, the client computational cost of VeriFL differs from competing protocols by additional $\frac{1}{E}$ factor. Notable, since the core idea of all approaches of HP-based category is very similar, we observe that this optimization could be also applied to any of them.

There are also two protocols that lie at the intersection of the RA- and HP-based categories: VerifyNet \cite{VerifyNet} and SVFL \cite{SVFL}. In VerifyNet, clients share five additional values along with their gradients, which are later aggregated by the server. Aggregated values are shared back with clients so that they can verify the correctness of the gradients aggregation relying on the properties of the homomorphic hash function. The second approach, SVFL, is based on a homomorphic signature scheme. During the training procedure, the server performs a  redundant aggregation of signatures and each client runs the verification algorithm based on the HP. Authors considered a malicious threat model for the server, however one should be careful while using signature-based verification techniques due to the concerns described above for similar approaches of the RA-based verification category \cite{SVeriFL, Zhangetal}. 

\textbf{ZKP based verification.} The third category contains approaches which are based on ZKPs. The core principle could be described as follows: a party performs calculations and at the same time computes the proof, which is shared along with the result of calculations; other parties can later run the proof verification algorithm to ensure that the result was computed correctly. In contrast to previous categories, advanced ZKPs allow proving arbitrary computations, therefore such methods are suitable for proving the correctness of both aggregation of clients' values \cite{zkFL, zkDFL, GOPA} and local computations \cite{Federify, Heissetal, Ruckeletal}. Moreover, recent ZKP schemes provide a proof size that is sublinear in the amount of computations to prove.

In protocols focused on aggregation, authors build on different infrastructures and ZKP schemes. In GOPA \cite{GOPA}, authors introduced a decentralized gossip approach where nodes publish proofs of their computations using $\Sigma$-protocols. In zkFL \cite{zkFL} authors apply more modern ZKP scheme, Halo2, and provide two versions of the protocol -- with a centralized FL setting and a blockchain based one. In zkDFL \cite{zkDFL} authors rely on the blockchain infrastructure and the Groth16 scheme. We also observed that authors of zkDFL and zkFL use different techniques to prove that each client value was indeed sent by one of the clients. In zkFL, all clients sign commitments to their local models and the server includes the signature verification in the proof of own computations.  Interestingly, with this example one can notice that ZKPs help to eliminate the need for clients to communicate with each other to verify the integrity of aggregated data. In zkDFL, authors proposed to deploy a smart contract which checks that the result of redundant aggregation of local weights hashes performed by server is equal to the sum of hashes uploaded by clients (notably, computational burden of the aggregation verification is also transferred to the smart contract). Consequently, differences in the settings and chosen ZKP schemes result in different complexity metrics. 

In contrast to protocols with a verifiable aggregation, approaches focused on local computations verification \cite{Ruckeletal, Heissetal, Federify} have almost identical design: all protocols use the Groth16 scheme to achieve verifiability of clients' local computations in a blockchain based setting. However, authors focus on verification of different ML models: a linear regression model in \cite{Ruckeletal}, a naive Bayes classifier in \cite{Federify} and a feedforward neural network in \cite{Heissetal}. In all three approaches, aggregation is performed by a smart contract, therefore the correctness of the aggregation relies on the blockchain security assumptions. One can see that proving, verification and proof size complexities would be the same for the same type of computations, since all three protocols use the same ZKP scheme. Similarly, in all three protocols participants need to have an access to public parameters (including a CRS) which would also require the same amount of memory for their storage. We note that Table \ref{VerLocModTable} does not reflect the complexities of miners which perform computations represented as smart contracts of the blockchain-based approaches.

Additionally, in \cite{Ruckeletal} clients commit to the Merkle root of their private dataset  while registering in the system which allows ensuring data integrity within the whole training procedure. To achieve this, clients extend each proof showing that computations are performed with committed data. In \cite{Federify}, ZKPs are also applied to prove properties of local models, i.e., the distance metrics indicating how close the submitted model is to the global one. 

Lastly, it is important to mention that there are many research works devoted to proving the correctness of ML calculations without a FL setting \cite{zkMLaaS, VeriML, VerifiableSupVecMachine}. Often, the same verification techniques and developed optimizations could be applied in the context of FL to prove local computations of participants. We believe that the vast majority of such techniques would fit the ZKP-based verification category of our taxonomy.

\textbf{Data Embedding based verification.} This category covers protocols, where participants embed additional values into their data before sharing it with untrusted parties; later, the result of calculations performed by an untrusted source is assumed to be correct if the corresponding additional values are computed correctly. The embedding principle leads to an increase in the size of the transmitted data and the complexity of the outsourced calculations, which depends on the size of embedded values, and require more expensive data preprocessing.

One of the best approaches focused on the verifiable aggregation from the  complexity perspective is VFL 
\cite{VFL}. In VFL clients encode their secret values as a polynomial function and embed an additional point $(a_i, A)$ before interpolation, where $a_i$ is a parameter of client $i$ and $A$ is obtained from a pseudo random generator (PRG) using $a_i$ as an input. In order to verify the result of aggregation performed by a malicious server, each client check that the evaluation of the aggregated polynomial function at the point $a_i$ corresponds to the output of PRG with $\sum_{i=1}^{C}a_i$ as an input. In this protocol, transmitted data overhead is negligible, but each client has to compute a costly interpolation and stores large public parameters. 

In PVD-FL \cite{PVD_FL}, authors proposed a decentralized verifiable protocol which is based on the verifiable matrix multiplication algorithm. Parties embed random vectors into their data and then check that these vectors were correctly multiplied, as a result, the developed algorithm allows to verify basic operations of ML. In contrast to other approaches surveyed in this paper, in PVD-FL authors aim to show correctness of both aggregation and local models calculation. However, they also mention that their protocol is still vulnerable against poisoning attacks.

\subsection{Discussion}

In this subsection, we discuss how various features of the cross-silo setting impact the development of a verifiable FL protocol. We highlight efficient schemes which cope with cross-silo FL challenges, describe the influence of the threat model choice on the efficiency and security of the protocol, and discuss advantages and disadvantages of the blockchain-based approaches.

Firstly, we describe the link between cross-silo setting characteristics and the efficiency of protocols. One can notice that in tables \ref{VerAggTable} and \ref{VerLocModTable} there are mainly two parameters determining communication cost: $C$ and $D$. However, there is a large difference in their impact on the complexity. Since the number of participants in cross-silo settings is moderate while ML models typically have large sizes, a dependence on $D$ is less desirable. Nevertheless, taking into account that clients anyway must send their local models to a server with $O(D)$ communication cost, the overall FL complexity would become asymptotically worse only in cases when the verification overhead is larger than $D$. For example, such as in \cite{DEVA, NIVA, FedTrust}, where the communication cost is $O(CD)$. 

\label{blockchain_discussion}
Secondly, we observe that several approaches rely on a blockchain infrastructure \cite{PPV-BC, BFLbasedBytoChain, zkDFL, Ruckeletal, Heissetal}. This strategy offers several advantages. For instance, smart contracts enforce a transparent and verifiable distribution of incentives \cite{Ruckeletal}. The use of smart contracts to perform aggregation also makes the presence of a distinct server unnecessary, thereby replacing a single party trust with blockchain trust guarantees. All blockchain-based approaches are also robust against limited drop-outs of aggregators, i.e. miners.  Table \ref{VerAggTable} demonstrates that the verification overhead per client is generally smaller for blockchain-based approaches compared to non-blockchain protocols with similar verification techniques; however, there is a significant infrastructure overhead, e.g., costs of the miners, which is a critical limitation  within the context of cross-silo FL.  The miners have to execute identical calculations, resulting in a tremendous total computational burden across all participants. We note that the miners' overhead can vary across different blockchains depending on the consensus mechanism: while in Proof-of-Work (PoW) systems like Bitcoin, all miners redundantly execute all computations, in Proof-of-Stake (PoS) systems like Ethereum, transactions are verified by a set of parties called "validators".  Although the general idea remains the same, the set of validators is smaller, reducing the overall computational cost. Finally, in cross-silo FL there is typically at least one party interested in obtaining the results of training, thus there is no a strong need for a decentralized infrastructure.

Thirdly, during our analysis of verifiable aggregation protocols, we observed that server has several ways to fabricate the aggregation result, and sometimes a proposed protocol only addresses a portion of them. Specifically, there are three ways in which a forger can manipulate the aggregation of clients' values:  omitting one of the values, replacing one update with arbitrary data or inserting an additional update. However, in RA-based approaches that rely on cryptographic signature schemes, a verifier is only capable of checking that server has not omitted values from other clients and has not inserted additional values in the sum. Nevertheless, a malicious server still may aggregate arbitrary signed values and successfully pass the verification. Therefore, such verification techniques should be employed with caution.

Almost all verifiable aggregation approaches from our analysis rely on honest or honest-but-curios clients' threat model. The only exceptions with support of malicious clients are: the blockchain based approach BytoChain \cite{BFLbasedBytoChain}, where a committee of verifiers is tasked to check properties of clients' uploads, therefore partially mitigating poisoning attacks and free-riding attacks.
 Indeed, weaker threat models are common in cross-silo FL. However, in real world conditions the use of such threat models is not advisable: a protocol that is resistant to malicious attacks incentivizes participants to avoid cheating, as opposed to a protocol designed with a weaker threat model.


Lastly, our observations indicate that the majority of verifiable aggregation protocols rely on either redundant aggregation mechanisms or leverage the homomoprhic properties of cryptographic primitives. However, these techniques are not suitable for proving complex computations, such as those encountered in ML. To address this limitation, researchers employ ZKP schemes, known for the ability to prove arbitrary computations. We examined several ZKP-based protocols developed both to verify aggregation and more complex computation of clients. Furthermore, ZKP schemes offer the flexibility to reinforce the primary proof with additional information, showing the correctness of computed noise, data provenance, or distance metrics. The complexity metrics of ZKP-based FL protocols directly depend on the chosen scheme, however we observe the lack of justifications behind the scheme selection process in the literature. Considering the benefits of ZKPs and their advantage in comparison with other techniques, in the next section we examine existing ZKP schemes with a focus on their applicability in the cross-silo setting.

\section{ZKP for cross-silo FL} \label{ZKPschemesAnalysis}

The development of new ZKP schemes has been an active area of research over the past decade. While the surge of new protocols has led to a broad variety of schemes to choose from, it also resulted in additional desirable characteristics, making it challenging to determine the most suitable choice for a specific application. In this context, we discuss the applicability of ZKPs in cross-silo FL and study how to prove calculations in this setting minimizing the cost.

\subsection{Applicability}

In this subsection, we discuss diverse characteristics of ZKP schemes, examining their implications within the context of the cross-silo FL setting. Specifically, we consider the time complexities for proving, verifying and preprocessing, the proof size, the common reference string (CRS) and the commit-and-prove property which some ZKP schemes feature, and the partition into transparent schemes versus trusted setup based schemes.

\textbf{Computational complexity of proving and verifying.}
In cross-silo FL, the verification of proofs generated during the training procedure presents a significant challenge: each party who wants to ensure the correctness of the protocol needs to execute a verification algorithm in order to check proofs coming from many participants. If the verification algorithm has a linear cost with respect to the size of the computations, the cost for the verifier is proportional to the combined computations of all other parties in the system, which is often too expensive. Furthermore, in scenarios where the protocol is publicly verifiable, e.g. when all proofs are stored on a bulletin board as demonstrated in \cite{GOPA, NIVA}, verification of all proofs performed by an external party after the training procedure would become excessively time-consuming. Hence, we consider it is desirable that the verification time complexity is at most logarithmic in the total amount of computation to ensure feasibility within the FL setting. In contrast to the verification complexity, the proving complexity is of a slightly less priority since each party has to prove only its own computations once. Consequently, the overhead incurred by executing an algorithm to construct a ZKP is often feasible even if this algorithm is somewhat more expensive.

\textbf{Preprocessing computational complexity.}
Achieving fast verification and small proofs often comes with a costly preprocessing phase, when the CRS is generated. Since this phase is completed only once before the training process, while the preprocessing compexity is a factor to consider, its impact is generally less critical than for algorithms that are executed within the training loop.

\textbf{Proof size.}
In a verifiable FL protocol, all parties compute proofs of their local calculations. If the proof size scales linearly with the size of computations, parties need to share data proportional to that computation. Such proofs would require gigabytes to petabytes of storage. In the context of publicly verifiable protocols, this problem becomes even more acute. Therefore, similarly to the requirement for verification cost, it is essential for the proof size to be at most logarithmic. 

\textbf{Common Reference Strings (CRS).} 
Numerous SNARKs rely on generating a CRS during the preprocessing phase. There exist two types of CRS: universal, capable of supporting all circuits of bounded size, and custom, applicable only for specific circuits. Given that in FL all parties agree on the model to train, i.e. computations to prove, there is no need to use a scheme with a universal reference string. 

In the context of FL, all participants have to access the CRS to compute and verify proofs. The cost of delivering the CRS is determined both by the size of this CRS, and the challenge to broadcast this CRS in a way such that all recipients can trust it. One should take into account that the complexity of the CRS's size differs depending on the ZKP scheme. Therefore, if the CRS scales linearly to the size of computations to prove and some heavy computations have to be proven, the storage cost will likely become prohibitively large.

\textbf{Transparent and trusted setup based setups.}
 A huge fraction of ZKP schemes depend on trusted setups to generate public parameters. Modern protocols often require only one honest party for this procedure. Since many cross-silo FL protocols are developed under the assumption that at least one honest party is present in the setting, applying such schemes would fit the threat model of these protocols. At the same time, there exist transparent ZKP schemes that eliminate the need for a trusted setup. As a result, integrating such schemes into FL protocol would allow reliance on a more robust threat model. 

\textbf{Commit-and-prove.}
Some ZKP schemes maintain the commit-and-prove property, meaning that
the prover can commit to certain data and later prove computations involving this data. As a result, a verifier will be able to check that data has not changed across multiple proofs. In the context of FL, this property is useful to demonstrate that DOs have not changed data during the training. Such a procedure mitigates the risk of one-shot data poisoning attacks and narrows capabilities of potential attackers.

Following the discussion above, in Table \ref{table:ZKPComparison}, we present an asymptotic comparison of various ZKP schemes with at most logarithmic size proofs realtive to the size of computations. Our comparison shows that while numerous schemes offer small proof sizes and fast verification, only DARK-based approaches (e.g., SuperSonic, Dew) maintain a constant size of public parameters at the same time. 

\begin{table*}
\begin{tabular}{ |c | c | c | c | c | }
\hline
\textbf{Scheme} & \textbf{Parameters size} & \textbf{Proving}& \textbf{Verification} & \textbf{Proof Size}  \\ \hline
Dory \cite{ZKP_Dory} & $O(|C|)$ & $O(|C|)$ & $O($log$|C|)$ & $O($log$|C|)$  \\
Gemini (space-efficient) \cite{ZKP_Gemini} & $O(|C|)$ &  $O(|C|$log$^{2}|C|)$ & $O($log$|C|)$ & $O($log$|C|)$  \\
Gemini (time-efficient) \cite{ZKP_Gemini} & $O(|C|)$ & $O(|C|)$ & $O($log$|C|)$ & $O($log$|C|)$  \\
SuperSonic \cite{ZKP_SuperSonic}, DARK-fix \cite{ZKP_Dew} & $O(1)$ & $O(|C|$log$|C|)$ & $O($log$|C|)$ & $O($log$|C|)$ \\ 

BCCGP \cite{ZKP_BCCGP} & $O(|C|)$ & $O(|C|)$ & $O(|C|)$ & $O($log$|C|)$ \\ 
Bulletproofs \cite{ZKP_Bulletproof} & $O(|C|)$ & $O(|C|)$ & $O(|C|)$ & $O($log$|C|)$ \\ 
\makecell{Compressed\\ $\Sigma$-protocol \cite{ZKP_CompressedSigma} } & $O(|C|)$ & $O(|C|)$ & $O(|N|)$ & $O($log$(|C|))$\\

Groth16 \cite{ZKP_Groth} & $O(|C|)$ & $O(|C|$log$|C|)$ & $O(|N|)$ & $O(1)$ \\ 
Sonic \cite{ZKP_Sonic} & $O(|C|)$ & $O(|C|$log$|C|)$  & $O(N)$ & $O(1)$ \\ 
GGPR \cite{ZKP_GGPR} & $O(|C|)$ & $O(|C|$log$|C|)$ & $O(|N|)$ & $O(1)$  \\ 
Pinochio \cite{ZKP_Pinochio} & $O(|C|)$ & $O(|C|$log$|C|)$ & $O(|N|)$ & $O(1)$  \\

PLONK \cite{ZKP_PLONK} & $O(|C|)$ & $O(|C|$log$|C|)$ & $O(|N|)$ & $O(1)$ \\
vnTinyRAM \cite{ZKP_vnTinyRam}  & $O(|C|$log$|C|)$ & $O(|C|$log$^{2}|C|)$ & $O(|N|)$ & $O(1)$ \\
Mirage \cite{ZKP_Mirage} & $O(|C|)$ & $O(|C|$log$|C|)$ & $O(|N|)$ & $O(1)$ \\
Behemoth \cite{ZKP_Behemoth} & $O(|C|)$ & $O(|C|^3$log$|C|)$ & $O(|N|)$ & $O(1)$ \\
Dew \cite{ZKP_Dew} & $O(1)$ & $O(|C|^2)$ & $O($log$|C|)$ & $O(1)$ \\ 
\hline

\end{tabular}
\caption{Asymptotic comparison of ZKP schemes with logarithmic and constant proof size complexity. $C$ is the computation expressed as a circuit, $|C|$ is the number of gates in the circuit, $|N|$ is the length of inputs.}
\label{table:ZKPComparison}
\end{table*}

\subsection{Storage cost optimization}
\label{subsection:packing}

In the previous subsection, we highlighted that one of the more important criteria for designing a verifiable FL protocol is the total communication cost of ZKP schemes and the associated cost to store the proofs. If parties perform more computations than they want to include in a single proof, they can distribute their computations over multiple proofs.  In this subsection we first study what is the best granularity of the proofs under different ZKP schemes and then discuss an alternative idea to prove computations based on recursive proof composition. 

As in a ML algorithm the same operations are often repeated many times, e.g., for different data or for different epochs in iterative algorithms, we assume that one can produce a ZKP for all computations by repeatedly proving correct evaluation of a single circuit. Then, we introduce a parameter $k$ representing how many evaluations of this circuit are included per ZKP. In order to understand the effect of grouping less or more computations together in a single proof, we define a function to compute the storage cost as a function of $k$ and then find the optimal number $k$ of grouped circuits to prove, i.e., we find the $k$ with minimum function value.

To begin with, we introduce the following notations:

\begin{itemize}
    \item $c \in  \mathbb{N} $ -- the size of the minimal circuit in bits, i.e., the size of the smallest circuit so that the complete algorithm can be represented as a repetition of that circuit;
    \item $n \in  \mathbb{N} $ -- the number of circuits to prove; 
    \item $1 \leq k \in  \mathbb{N}  \leq n$ -- the number of circuits grouped for one proof;
    \item $\delta : \mathbb{N}^3 \rightarrow \mathbb{N}$ -- the function that, for a certain ZKP scheme, gives the total communication cost $\delta(k; c,n)$ in bits to prove an FL algorithm consisting of $n$ evaluations of a circuit of size $c$, where proofs are given for groups of $k$ circuits;
    \item $ \lambda: \mathbb{N} \rightarrow \mathbb{N} $ -- the function that returns the proof size $\lambda(c)$ required to prove a circuit of size $c$ in bits for a certain ZKP scheme;
    \item $\psi: \mathbb{N} \rightarrow \mathbb{N}$ -- the function that gives the size of the public parameters $\psi(c)$ required to prove circuits of size $c$ in bits for a certain ZKP scheme.
\end{itemize}
We also note that the the cost per bit for the distribution and storage of proofs may be different than the cost per bit for the generation and distribution of public parameters, hence we scale $\psi$ and $\lambda$ with constants $a$ and $b$ to reflect the appropriate weights of a certain ZKP scheme.

Then, for an arbitrary ZKP scheme we can define $\delta$ in the following way: 

\begin{equation}
\delta(k; c, n) = a\psi(kc) +  \frac{nb \lambda(kc)}{k}    
\end{equation}

One can see that if
$k=1$, then the storage cost is
as in a classic verifiable FL protocol without the storage optimization. 

One can notice that in Table \ref{table:ZKPComparison} there are ZKP schemes with the same asymptotic complexities of proof and parameters size. Since in the scope of this optimization, such schemes have the same optimal value of $k$, we group schemes with the same complexities together and describe each group below. 

\textbf{Group 1: Constant parameters size and constant proof size.}

We define the function $\delta_1$ by instantiating $\psi$ and $\lambda$ with corresponding asymptotic complexities of the Group 1 in the definition of $\delta$: 

\begin{gather*}
\delta_1(k; c, n) = a +  \frac{nb}{k} \\
\end{gather*}

One can see that in the simple case when a ZKP scheme (for example, the Dew scheme \cite{ZKP_Dew}) has constant parameters and proof size, the function tends to a minimum with the increase of $k$, thus the optimal $k=n$.

\textbf{Group 2: Constant parameters size and logarithmic proof size.}

Several approaches of this group can be found in Table \ref{table:ZKPComparison}:
SuperSonic \cite{ZKP_SuperSonic} and its improved version called DARK-fix \cite{ZKP_Dew}. In order to find an optimal $k$ for this group, we define $\delta_2$ similarly to the $\delta_1$ using the logarithmic proof size complexity: 

\begin{gather*}
  \delta_2(k; c, n) = a +  \frac{nb \cdot log(ck)}{k} \\
\end{gather*}

Since for k, c > 1 the function is decreasing, the minimum is reached at $k=n$.

\textbf{Group 3: Linear parameters size and constant proof size.}
In table \ref{table:ZKPComparison}, there are many approaches with characteristics of the Group 3: Groth16 \cite{ZKP_Groth}, Sonic \cite{ZKP_Sonic}, GGPR \cite{ZKP_GGPR}, Pinochio \cite{ZKP_Pinochio}, PLONK \cite{ZKP_PLONK}, Mirage \cite{ZKP_Mirage}, Behemoth \cite{ZKP_Behemoth}. We define $\delta_3$ as follows:

\begin{gather*}
\delta_3(k; c, n) = akc +  \frac{nb}{k}
\end{gather*}

We can approximately find the minimum of this function on integers by taking the derivative of the corresponding function on the reals:

\begin{gather*}
\frac{ d\delta_3(k; c, n)}{dk} = ac - \frac{nb}{k^2}
\end{gather*}

The minimum value of $\delta_3$ is $k_e = \sqrt{\frac{nb}{ac}}$. Since $k_e$ could be less than one, the optimal $k$ is $min(max(k_e, 1), n)$. Note, that in practice one has to check which of the two values $\lfloor k_e\rfloor$ or $\lceil k_e\rceil$ would give a minimal value of $\delta_3$.

\textbf{Group 4: Linear parameters size and logarithmic proof size.}
Schemes which correspond to the Group 4 are: Dory \cite{ZKP_Dory}, Gemini \cite{ZKP_Gemini}, Bulletproofs \cite{ZKP_Bulletproof} and  Compressed Sigma-Protocol \cite{ZKP_CompressedSigma}. Taking into account their complexities, we define $\delta_4$ in the following way:

\begin{gather*}
\delta_4(k; c, n) = ack + \frac{nb \cdot log(ck)}{k} \\
\end{gather*}

Let $r = \frac{a}{bnc}$ and $K = kc$, then 

\begin{gather*}
\delta_4(k; c, n) = bnc( rK + \frac{log(K)}{K}) \\
\end{gather*}

\begin{figure*}[ht]
\centering
 \includegraphics[scale=0.6]{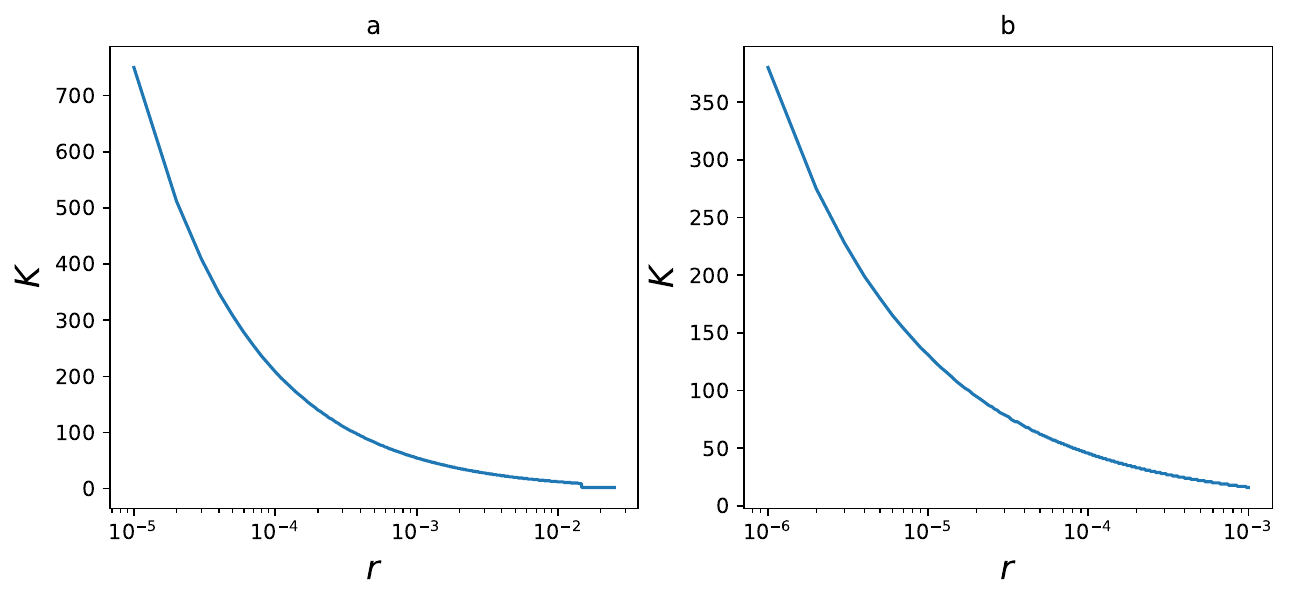}
\caption{The dependence of the optimal $K$ on $r$ for $\delta_4$ (a) and $\delta_5$ (b).}
\label{FigureOptimalK}
\end{figure*}

One can notice that the optimal value of $K$ depends on the value of $r$: the smaller the value of $r$, the larger the value of the optimal $K$. We present a plot demonstrating the dependence of the optimal $K$ on $r$ in Figure \ref{FigureOptimalK} (a). As a result, using this plot, one can choose the ZKP scheme with linear parameters size and logarithmic proof size and infer the optimal value of $k$ with respect to the parameters of the chosen scheme and parameters of the FL setting.

\textbf{Group 5: |C|log|C| parameters size and constant proof size.}
Lastly, we describe $\delta_5$ which corresponds to the Group 5 which consist of only one approach: vnTinyRAM \cite{ZKP_vnTinyRam}.

\begin{gather*}
\delta_5(k; c, n) = ack \cdot log(ck) + \frac{nb}{k} \\
\end{gather*}

Applying the same substitution as for the $\delta_4$: 

\begin{gather*}
\delta_5(k; c, n) = bnc(rK \cdot log(K) + \frac{1}{K}) \\
\end{gather*}

Similarly, to the previous group, the plot demonstrating the dependence of the optimal $K$ on $r$ is presented in Figure \ref{FigureOptimalK} (b).

With the above generally applicable strategy, one can determine the optimal number of grouped circuits to prove for various ZKP schemes in order to minimize the storage cost or, in other words, communication cost of a verifiable ZKP-based FL protocol. 

Still, once the number of circuits in a group is fixed, the costs get linear in the number of groups.  In practice, while the simple strategy discussed above already allows one to handle reasonably large learning tasks, 
there is a maximal size to a group, e.g., due to memory storage limits.  Hence, the asymptotic cost given a bound on memory storage remains linear, and given the ever growing size of ML tasks one could wonder whether one can do even better.

One idea which may address this issue, but has not been tried in practical FL settings, 
  is to compose proofs recursively, in particular to use a ZKP scheme that allows for constructing proofs that attest to the correctness of both circuits and other proofs in the same ZKP scheme. 
  Following this idea, at the beginning a prover creates a proof for the first part of computations and then iteratively generates a proof for the correctness of both the next circuit and all preceding proofs.
If successful, this idea then could keep the work of the prover about linear in the number of computations, the memory storage of the prover constant, the proof size and communication cost constant and the verifier cost logarithmic.  


Several papers have studied the theory of this type of approach.
For example, in the context of incrementally verifiable computations (IVC) \cite{ZKP_Nova, ZKP_SuperNova, ZKP_HyperNova, ZKP_ProtoStar}, proof-carrying data (PCD) \cite{PCD_folding} and zk Virtual Machine (zkVM) \cite{zkVMCeno}. State-of-the-art solutions mostly rely on a folding scheme, a primitive that reduces the task of checking two instances of some relation to the task of checking a single instance \cite{ZKP_Nova}. For instance, authors of Nova \cite{ZKP_Nova} were the first to apply the folding scheme to IVC, resulting in a scheme where the prover's work is linear to the size of the smallest circuit, and the verifier's time and the proof size are logarithmic. Later, other authors presented advanced versions of the protocol such as SuperNova \cite{ZKP_SuperNova}, HyperNova \cite{ZKP_HyperNova}, and ProtoStar \cite{ZKP_ProtoStar}, extending the original approach to more general settings and more efficient arithmetizations.

However, to the best of our knowledge, there are neither works studying an efficient recursive proof composition to verify computations in FL, nor works exploring whether solutions developed for IVC, PCD and zk VM could be adapted to FL. Among all protocols analyzed in Section \ref{ApproachesAnalysis}, only in zkFL \cite{zkFL} authors apply a recursive ZKP scheme: Halo 2. Nevertheless, they do not discuss the construction of a minimal circuit or recursion properties of the chosen ZKP and Halo 2 is an early recursive ZKP scheme with linear verifier cost.  A linear verifier cost is not practical for large ML tasks.  Therefore, the impact of recursion possibilities of the scheme on the FL complexities remains unclear.

We believe that applying more efficient recursive proof composition to verify FL computation remains an open research question and can potentially reduce the gap that prevents the use of ZKPs in practical applications.


\section{Challenges and future directions} \label{ChallengesFutureDirections}

While verifiable cross-silo FL protocols is a well studied area with a plenty of solutions, our analysis reveals several challenges which have not been yet addressed by the research community. Below we discuss potential research gaps and future directions.

Firstly, to the best of our knowledge, there are currently no verifiable FL protocols that fully support verification of both computations performed by clients and server at the same time. Exceptionally, PVD-FL \cite{PVD_FL} aims to achieve this, however the approach does not guarantee the correctness of all actions of participants and is still vulnerable to poisoning attacks. While some papers, such as \cite{BFLbasedBytoChain}, consider threats from both the server and clients, authors do not verify the clients' behavior, but analyze the distribution of their values to address potential attacks. Instead, we believe it would be an interesting direction to design a protocol that would enable verification (following the definition in \ref{VerifFLDefinition}) of actions of all participants. It would be later intriguing to compare different approaches from the threat models and the efficiency perspectives.

Secondly, we observe that verifiable aggregation is primarily studied for the most popular type of aggregation -- averaging of vectors possessed by DOs. Nevertheless, in certain settings, other U-statistics with a kernel of the degree two or larger (e.g. Kendall rank correlation coefficient) could be applied \cite{U-Statistics}, introducing new challenges in the verification process. Based on our analysis, we can see that ZKP-based verification would potentially fit, however finding an efficient solution still remains an open problem.

Thirdly, among the papers considered in this SOK,
there are no protocols that are robust against a collusion between client and server to bypass the verification. However, in real world scenarios such collusion might occur. Interestingly, in VerSA \cite{VerSA}, the authors do not consider collusion attacks, justifying this choice with the impossibility result shown by Gordon et al. \cite{CollusionImpossible}. The authors of the latter paper demonstrated that in specific settings multiclient verifiable computation cannot be achieved in the presence of users colluding with the server. Nonetheless, the specific setting considered does not necessarily apply to FL. For instance, in contrast to the setting from \cite{CollusionImpossible}, in a verifiable FL protocol clients may exchange messages with each other or prove computations interactively. As a result, this impossibility result does not necessarily show that a verifiable FL protocol cannot be developed to be robust against server-client collusion attacks. We believe that the development of such a protocol is an interesting direction for future research.

Fourthly, we observe that the repetitive nature of FL training is usually overlooked while developing a verifiable protocol. Nevertheless, this property opens up an opportunity to design various optimizations. For instance, in \cite{VeriFL}, the authors considered this property to combine verification of multiple epochs together, thereby reducing the computational cost of clients. Following similar ideas, in Subsection \ref{subsection:packing}, we proposed an optimization based on the observation that instead of repetitive proofs in FL, one can group circuits together before proving, thereby optimizing communication costs. However, we believe that other optimizations, for example, for approaches from other taxonomy groups, require further analysis.   

Lastly, as described in Subsection \ref{subsection:packing}, to the best of our knowledge, there are no works exploring the applicability of recursive ZKP schemes in the context of FL. In recent years, there has been an active research in the ZKP community to develop such schemes \cite{ZKP_Halo, ZKP_Nova, ZKP_HyperNova}. We anticipate that they deserve a particular attention. Their characteristics may allow for new optimizations and significant reduction in complexities of verifiable protocols.

\section{Conclusion}

In this paper, we presented a systematization of knowledge on verifiable cross-silo FL. We proposed a new taxonomy distinguishing four categories of verification techniques. We described general design patterns for each category and provided an analysis of threat models, computational and communication costs both per client and per server for each protocol. We also discussed how various features of the cross-silo setting impact the verification process and highlighted advantages of ZKP-based protocols. As a continuation of our conclusions, we discussed the applicability of different ZKP schemes for cross-silo FL and optimization strategies to minimize the communication cost. Finally, we described several research challenges revealed in our analysis and indicated future scientific directions.

\section{Acknowledgments}

This project was partially supported by the ’Chair TIP’ project funded by ANR, I-SITE, INRIA and MEL, and the Horizon Europe TRUMPET project grant no. 101070038.

\bibliographystyle{abbrv}
\bibliography{references}

\begin{thebibliography}{10}

\bibitem{zkDFL}
M.~Ahmadi and R.~Nourmohammadi.
\newblock zkdfl: An efficient and privacy-preserving decentralized federated
  learning with zero-knowledge proof.
\newblock
  \url{https://synthical.com/article/a8c00457-dadd-4207-bd23-7edaf0188617}, 11
  2023.

\bibitem{ZKP_Dew}
A.~Arun, C.~Ganesh, S.~Lokam, T.~Mopuri, and S.~Sridhar.
\newblock Dew: A transparent constant-sized polynomial commitment scheme.
\newblock In {\em Public-Key Cryptography – PKC 2023: 26th IACR International
  Conference on Practice and Theory of Public-Key Cryptography, Atlanta, GA,
  USA, May 7–10, 2023, Proceedings, Part II}, page 542–571, Berlin,
  Heidelberg, 2023. Springer-Verlag.

\bibitem{ZKP_CompressedSigma}
T.~Attema and R.~Cramer.
\newblock Compressed $\sigma$-protocol theory and practical application to plug
  {\&} play secure algorithmics.
\newblock In D.~Micciancio and T.~Ristenpart, editors, {\em Advances in
  Cryptology -- CRYPTO 2020}, pages 513--543, Cham, 2020. Springer
  International Publishing.

\bibitem{U-Statistics}
J.~Bell, A.~Bellet, A.~Gascon, and T.~Kulkarni.
\newblock Private protocols for u-statistics in the local model and beyond.
\newblock In S.~Chiappa and R.~Calandra, editors, {\em Proceedings of the
  Twenty Third International Conference on Artificial Intelligence and
  Statistics}, volume 108 of {\em Proceedings of Machine Learning Research},
  pages 1573--1583, Online, 26--28 Aug 2020. PMLR.

\bibitem{ZKP_vnTinyRam}
E.~Ben-Sasson, A.~Chiesa, E.~Tromer, and M.~Virza.
\newblock Succinct {Non-Interactive} zero knowledge for a von neumann
  architecture.
\newblock In {\em 23rd USENIX Security Symposium (USENIX Security 14)}, pages
  781--796, San Diego, CA, Aug. 2014. USENIX Association.

\bibitem{FirstCS}
M.~Blum.
\newblock Coin flipping by telephone a protocol for solving impossible
  problems.
\newblock {\em SIGACT News}, 15(1):23–27, jan 1983.

\bibitem{ZKP_BCCGP}
J.~Bootle, A.~Cerulli, P.~Chaidos, J.~Groth, and C.~Petit.
\newblock Efficient zero-knowledge arguments for arithmetic circuits in the
  discrete log setting.
\newblock In {\em Proceedings, Part II, of the 35th Annual International
  Conference on Advances in Cryptology --- EUROCRYPT 2016 - Volume 9666}, page
  327–357, Berlin, Heidelberg, 2016. Springer-Verlag.

\bibitem{ZKP_Gemini}
J.~Bootle, A.~Chiesa, Y.~Hu, and M.~Orr{\'u}.
\newblock Gemini: Elastic snarks for diverse environments.
\newblock In O.~Dunkelman and S.~Dziembowski, editors, {\em Advances in
  Cryptology -- EUROCRYPT 2022}, pages 427--457, Cham, 2022. Springer
  International Publishing.

\bibitem{ZKP_Halo}
S.~Bowe, J.~Grigg, and D.~Hopwood.
\newblock Halo: Recursive proof composition without a trusted setup.
\newblock {\em IACR Cryptol. ePrint Arch.}, 2019:1021, 2019.

\bibitem{NIVA}
C.~Brunetta, G.~Tsaloli, B.~Liang, G.~Banegas, and A.~Mitrokotsa.
\newblock Non-interactive, secure verifiable aggregation for decentralized,
  privacy-preserving learning.
\newblock Cryptology ePrint Archive, Paper 2021/654, 2021.

\bibitem{ZKP_Bulletproof}
B.~Bunz, J.~Bootle, D.~Boneh, A.~Poelstra, P.~Wuille, and G.~Maxwell.
\newblock Bulletproofs: Short proofs for confidential transactions and more.
\newblock In {\em 2018 IEEE Symposium on Security and Privacy (SP)}, pages
  315--334, Los Alamitos, CA, USA, may 2018. IEEE Computer Society.

\bibitem{ZKP_SuperSonic}
B.~B{\"u}nz, B.~Fisch, and A.~Szepieniec.
\newblock Transparent snarks from dark compilers.
\newblock In A.~Canteaut and Y.~Ishai, editors, {\em Advances in Cryptology --
  EUROCRYPT 2020}, pages 677--706, Cham, 2020. Springer International
  Publishing.

\bibitem{ZKP_ProtoStar}
B.~Bünz and B.~Chen.
\newblock {ProtoStar}: Generic efficient accumulation/folding for special sound
  protocols.
\newblock Cryptology ePrint Archive, Paper 2023/620, 2023.
\newblock \url{https://eprint.iacr.org/2023/620}.

\bibitem{PPV-BC}
C.~Fang, Y.~Guo, J.~Ma, H.~Xie, and Y.~Wang.
\newblock A privacy-preserving and verifiable federated learning method based
  on blockchain.
\newblock {\em Computer Communications}, 186:1--11, 2022.

\bibitem{FiatShamir}
A.~Fiat and A.~Shamir.
\newblock How to prove yourself: Practical solutions to identification and
  signature problems.
\newblock In A.~M. Odlyzko, editor, {\em Advances in Cryptology --- CRYPTO'
  86}, pages 186--194, Berlin, Heidelberg, 1987. Springer Berlin Heidelberg.

\bibitem{VFL}
A.~Fu, X.~Zhang, N.~Xiong, Y.~Gao, H.~Wang, and J.~Zhang.
\newblock Vfl: A verifiable federated learning with privacy-preserving for big
  data in industrial iot.
\newblock {\em IEEE Transactions on Industrial Informatics}, 18(5):3316--3326,
  2022.

\bibitem{ZKP_PLONK}
A.~Gabizon, Z.~J. Williamson, and O.-M. Ciobotaru.
\newblock Plonk: Permutations over lagrange-bases for oecumenical
  noninteractive arguments of knowledge.
\newblock {\em IACR Cryptol. ePrint Arch.}, 2019:953, 2019.

\bibitem{SVeriFL}
H.~Gao, N.~He, and T.~Gao.
\newblock Sverifl: Successive verifiable federated learning with
  privacy-preserving.
\newblock {\em Information Sciences}, 622:98--114, 2023.

\bibitem{MedicineMLSurvey}
A.~Garg and V.~Mago.
\newblock Role of machine learning in medical research: A survey.
\newblock {\em Computer Science Review}, 40:100370, 2021.

\bibitem{ZKP_GGPR}
R.~Gennaro, C.~Gentry, B.~Parno, and M.~Raykova.
\newblock Quadratic span programs and succinct nizks without pcps.
\newblock In T.~Johansson and P.~Q. Nguyen, editors, {\em Advances in
  Cryptology -- EUROCRYPT 2013}, pages 626--645, Berlin, Heidelberg, 2013.
  Springer Berlin Heidelberg.

\bibitem{FirstZKP}
S.~Goldwasser, S.~Micali, and C.~Rackoff.
\newblock The knowledge complexity of interactive proof systems.
\newblock {\em SIAM Journal on Computing}, 18(1):186--208, 1989.

\bibitem{CollusionImpossible}
S.~D. Gordon, J.~Katz, F.-H. Liu, E.~Shi, and H.-S. Zhou.
\newblock Multi-client verifiable computation with stronger security
  guarantees.
\newblock In Y.~Dodis and J.~B. Nielsen, editors, {\em Theory of Cryptography},
  pages 144--168, Berlin, Heidelberg, 2015. Springer Berlin Heidelberg.

\bibitem{ZKP_Groth}
J.~Groth.
\newblock On the size of pairing-based non-interactive arguments.
\newblock In {\em Proceedings, Part II, of the 35th Annual International
  Conference on Advances in Cryptology --- EUROCRYPT 2016 - Volume 9666}, page
  305–326, Berlin, Heidelberg, 2016. Springer-Verlag.

\bibitem{VeriFL}
X.~Guo, Z.~Liu, J.~Li, J.~Gao, B.~Hou, C.~Dong, and T.~Baker.
\newblock Verifl: Communication-efficient and fast verifiable aggregation for
  federated learning.
\newblock {\em IEEE Transactions on Information Forensics and Security},
  16:1736--1751, 2021.

\bibitem{VerSA}
C.~Hahn, H.~Kim, M.~Kim, and J.~Hur.
\newblock Versa: Verifiable secure aggregation for cross-device federated
  learning.
\newblock {\em IEEE Transactions on Dependable and Secure Computing},
  20(1):36--52, 2023.

\bibitem{Heissetal}
J.~Heiss, E.~Grunewald, S.~Tai, N.~Haimerl, and S.~Schulte.
\newblock Advancing blockchain-based federated learning through verifiable
  off-chain computations.
\newblock In {\em 2022 IEEE International Conference on Blockchain
  (Blockchain)}, pages 194--201, Los Alamitos, CA, USA, aug 2022. IEEE Computer
  Society.

\bibitem{MarketingMLSurvey}
D.~Herhausen, S.~F. Bernritter, E.~W. Ngai, A.~Kumar, and D.~Delen.
\newblock Machine learning in marketing: Recent progress and future research
  directions.
\newblock {\em Journal of Business Research}, 170:114254, 2024.

\bibitem{FedTrust}
C.-F. Hsu, J.-L. Huang, F.-H. Liu, M.-C. Chang, and W.-C. Chen.
\newblock Fedtrust: Towards building secure robust and trustworthy moderators
  for federated learning.
\newblock In {\em 2022 IEEE 5th International Conference on Multimedia
  Information Processing and Retrieval (MIPR)}, pages 318--323, CA, USA, Aug.
  2022. IEEE.

\bibitem{VerifiableSupVecMachine}
C.~Hu, C.~Zhang, D.~Lei, T.~Wu, X.~Liu, and L.~Zhu.
\newblock Achieving privacy-preserving and verifiable support vector machine
  training in the cloud.
\newblock {\em IEEE Transactions on Information Forensics and Security},
  18:3476--3491, 2023.

\bibitem{CSFLSurvey}
C.~Huang, J.~Huang, and X.~Liu.
\newblock Cross-silo federated learning: Challenges and opportunities, 2022.

\bibitem{zkMLaaS}
C.~Huang, J.~Wang, H.~Chen, S.~Si, Z.~Huang, and J.~Xiao.
\newblock zkmlaas: a verifiable scheme for machine learning as a service.
\newblock In {\em GLOBECOM 2022 - 2022 IEEE Global Communications Conference},
  pages 5475--5480, Rio de Janeiro, Brazil, 2022. IEEE.

\bibitem{Federify}
G.~Keshavarzkalhori, C.~Pérez-Solà, G.~Navarro-Arribas,
  J.~Herrera-Joancomartí, and H.~Yajam.
\newblock Federify: A verifiable federated learning scheme based on zksnarks
  and blockchain.
\newblock {\em IEEE Access}, 12:3240--3255, 2024.

\bibitem{ZKP_Mirage}
A.~Kosba, D.~Papadopoulos, C.~Papamanthou, and D.~Song.
\newblock Mirage: Succinct arguments for randomized algorithms with
  applications to universal zk-snarks.
\newblock In {\em 29th USENIX Security Symposium}, pages 2129--2146, Boston,
  MA, USA, 08 2020. USENIX Association.

\bibitem{ZKP_SuperNova}
A.~Kothapalli and S.~Setty.
\newblock {SuperNova}: Proving universal machine executions without universal
  circuits.
\newblock Cryptology ePrint Archive, Paper 2022/1758, 2022.
\newblock \url{https://eprint.iacr.org/2022/1758}.

\bibitem{ZKP_HyperNova}
A.~Kothapalli and S.~Setty.
\newblock Hypernova: Recursive arguments for customizable constraint systems.
\newblock Cryptology ePrint Archive, Paper 2023/573, 2023.
\newblock \url{https://eprint.iacr.org/2023/573}.

\bibitem{ZKP_Nova}
A.~Kothapalli, S.~Setty, and I.~Tzialla.
\newblock Nova: Recursive zero-knowledge arguments from folding schemes.
\newblock In Y.~Dodis and T.~Shrimpton, editors, {\em Advances in Cryptology --
  CRYPTO 2022}, pages 359--388, Cham, 2022. Springer Nature Switzerland.

\bibitem{ModulusSurvey}
M.~Labs.
\newblock Zero-knowledge proof meets machine learning in verifiability: A
  survey.
\newblock
  \url{https://drive.google.com/file/d/1tylpowpaqcOhKQtYolPlqvx6R2Gv4IzE/view},
  2023.

\bibitem{ZKP_Dory}
J.~Lee.
\newblock Dory: Efficient, transparent arguments for generalised inner products
  and polynomial commitments.
\newblock In K.~Nissim and B.~Waters, editors, {\em Theory of Cryptography},
  pages 1--34, Cham, 2021. Springer International Publishing.

\bibitem{BFLbasedBytoChain}
Z.~Li, H.~Yu, T.~Zhou, L.~Luo, M.~Fan, Z.~Xu, and G.~Sun.
\newblock Byzantine resistant secure blockchained federated learning at the
  edge.
\newblock {\em IEEE Network}, 35(4):295--301, 2021.

\bibitem{zkVMCeno}
T.~Liu, Z.~Zhang, Y.~Zhang, W.~Hu, and Y.~Zhang.
\newblock Ceno: Non-uniform, segment and parallel zero-knowledge virtual
  machine.
\newblock Cryptology ePrint Archive, Paper 2024/387, 2024.
\newblock \url{https://eprint.iacr.org/2024/387}.

\bibitem{SVFL}
F.~Luo, S.~Al-Kuwari, and Y.~Ding.
\newblock Svfl: Efficient secure aggregation and verification for cross-silo
  federated learning.
\newblock {\em IEEE Transactions on Mobile Computing}, 23(1):850--864, 2024.

\bibitem{SFCV-FL}
A.~Madi, O.~Stan, A.~Mayoue, A.~Grivet-Sébert, C.~Gouy-Pailler, and R.~Sirdey.
\newblock A secure federated learning framework using homomorphic encryption
  and verifiable computing.
\newblock In {\em 2021 Reconciling Data Analytics, Automation, Privacy, and
  Security: A Big Data Challenge (RDAAPS)}, pages 1--8, McMaster University,
  Hamilton ON, Canada, 2021. IEEE.

\bibitem{ZKP_Sonic}
M.~Maller, S.~Bowe, M.~Kohlweiss, and S.~Meiklejohn.
\newblock Sonic: Zero-knowledge snarks from linear-size universal and updatable
  structured reference strings.
\newblock In {\em Proceedings of the 2019 ACM SIGSAC Conference on Computer and
  Communications Security}, CCS '19, page 2111–2128, New York, NY, USA, 2019.
  Association for Computing Machinery.

\bibitem{SAPSOK}
M.~Mohamad, M.~Önen, W.~Ben~Jaballah, and M.~Conti.
\newblock Sok: Secure aggregation based on cryptographic schemes for federated
  learning.
\newblock In IACR, editor, {\em PETS 2023, 23rd Privacy Enhancing Technologies
  Symposium, 10-15 July 2023, Lausanne, Switzerland (Hybrid Conference)}, pages
  140--157, Lausanne, 2023. Rochester NY: Privacy Enhancing Technologies Board.
\newblock IACR.

\bibitem{IntroZKP}
A.~Nitulescu.
\newblock zk-snarks: A gentle introduction.
\newblock Technical report, Ecole Normale Superieure, 2019.

\bibitem{FinanceMLSurvey}
A.~M. Ozbayoglu, M.~U. Gudelek, and O.~B. Sezer.
\newblock Deep learning for financial applications : A survey.
\newblock {\em Applied Soft Computing}, 93:106384, 2020.

\bibitem{ZKP_Pinochio}
B.~Parno, J.~Howell, C.~Gentry, and M.~Raykova.
\newblock Pinocchio: Nearly practical verifiable computation.
\newblock In {\em 2013 IEEE Symposium on Security and Privacy}, pages 238--252,
  San Francisco, California, US, 2013. IEEE.

\bibitem{EducationMLSurvey}
A.~S. Pinto, A.~Abreu, E.~Costa, and J.~Paiva.
\newblock How machine learning (ml) is transforming higher education: A
  systematic literature review.
\newblock {\em Journal of Information Systems Engineering and Management},
  8(2):21168, Apr. 2023.

\bibitem{FLattacksSurvey}
N.~Rodríguez-Barroso, D.~Jiménez-López, M.~V. Luzón, F.~Herrera, and
  E.~Martínez-Cámara.
\newblock Survey on federated learning threats: Concepts, taxonomy on attacks
  and defences, experimental study and challenges.
\newblock {\em Information Fusion}, 90:148--173, 2023.

\bibitem{Ruckeletal}
T.~Rückel, J.~Sedlmeir, and P.~Hofmann.
\newblock Fairness, integrity, and privacy in a scalable blockchain-based
  federated learning system.
\newblock {\em Computer Networks}, 202:108621, 2022.

\bibitem{GOPA}
C.~Sabater, A.~Bellet, and J.~Ramon.
\newblock An accurate, scalable and verifiable protocol for federated
  differentially private averaging.
\newblock {\em Machine Learning}, 111, 10 2022.

\bibitem{ZKP_Behemoth}
I.~A. Seres and P.~Burcsi.
\newblock Behemoth: transparent polynomial commitment scheme with constant
  opening proof size and verifier time.
\newblock Cryptology ePrint Archive, Paper 2023/670, 2023.
\newblock \url{https://eprint.iacr.org/2023/670}.

\bibitem{TrustworthyFL}
A.~Tariq, M.~A. Serhani, F.~Sallabi, T.~Qayyum, E.~S. Barka, and K.~A. Shuaib.
\newblock Trustworthy federated learning: A survey, 2023.

\bibitem{DEVA}
G.~Tsaloli, B.~Liang, C.~Brunetta, G.~Banegas, and A.~Mitrokotsa.
\newblock Deva: Decentralized, verifiable secure aggregation for
  privacy-preserving learning.
\newblock In {\em Information Security: 24th International Conference, ISC
  2021, Virtual Event, November 10–12, 2021, Proceedings}, page 296–319,
  Berlin, Heidelberg, 2021. Springer-Verlag.

\bibitem{BC_definition}
H.~Wang, Z.~Zheng, S.~Xie, H.-N. Dai, and X.~Chen.
\newblock Blockchain challenges and opportunities: a survey.
\newblock {\em International Journal of Web and Grid Services}, 14:352 -- 375,
  10 2018.

\bibitem{zkFL}
Z.~Wang, N.~Dong, J.~Sun, W.~Knottenbelt, and Y.~Guo.
\newblock zkfl: Zero-knowledge proof-based gradient aggregation for federated
  learning.
\newblock {\em IEEE Transactions on Big Data}, PP(01):1--14, may 2024.

\bibitem{ZKPmeetsML}
Z.~Xing, Z.~Zhang, J.~Liu, Z.~Zhang, M.~Li, L.~Zhu, and G.~Russello.
\newblock Zero-knowledge proof meets machine learning in verifiability: A
  survey, 2023.

\bibitem{VerifyNet}
G.~Xu, H.~Li, S.~Liu, K.~Yang, and X.~Lin.
\newblock Verifynet: Secure and verifiable federated learning.
\newblock {\em IEEE Transactions on Information Forensics and Security},
  15:911--926, 2020.

\bibitem{Zhangetal}
X.~Zhang, A.~Fu, H.~Wang, C.~Zhou, and Z.~Chen.
\newblock A privacy-preserving and verifiable federated learning scheme.
\newblock In {\em ICC 2020 - 2020 IEEE International Conference on
  Communications (ICC)}, pages 1--6, Online, 2020. IEEE.

\bibitem{TowardsVerifiableFL}
Y.~Zhang and H.~Yu.
\newblock Towards verifiable federated learning.
\newblock In {\em International Joint Conference on Artificial Intelligence},
  pages 5686--5693, Vienna, Austria, 2022. IJCAI.

\bibitem{PVD_FL}
J.~Zhao, H.~Zhu, F.~Wang, R.~Lu, Z.~Liu, and H.~Li.
\newblock Pvd-fl: A privacy-preserving and verifiable decentralized federated
  learning framework.
\newblock {\em IEEE Transactions on Information Forensics and Security},
  17:2059--2073, 2022.

\bibitem{VeriML}
L.~Zhao, Q.~Wang, C.~Wang, Q.~Li, C.~Shen, and B.~Feng.
\newblock Veriml: Enabling integrity assurances and fair payments for machine
  learning as a service.
\newblock {\em IEEE Transactions on Parallel and Distributed Systems},
  32(10):2524--2540, 2021.

\bibitem{PCD_folding}
Z.~Zhou, Z.~Zhang, Z.~Zhang, and J.~Dong.
\newblock Proof-carrying data from multi-folding schemes.
\newblock Cryptology {ePrint} Archive, Paper 2023/1282, 2023.

\bibitem{BC_smartcontracts}
W.~Zou, D.~Lo, P.~S. Kochhar, X.-B.~D. Le, X.~Xia, Y.~Feng, Z.~Chen, and B.~Xu.
\newblock Smart contract development: Challenges and opportunities.
\newblock {\em IEEE Transactions on Software Engineering}, 47(10):2084--2106,
  2021.

\end{thebibliography}

\end{document}